%% file: acl_latex.tex
\pdfoutput=1

\documentclass[11pt]{article}

\usepackage[preprint]{acl}

\usepackage{times}
\usepackage{latexsym}

\usepackage[T1]{fontenc}

\usepackage[utf8]{inputenc}

\usepackage{microtype}

\usepackage{inconsolata}

\usepackage{graphicx}

\input{math-commands}

\usepackage[utf8]{inputenc} %
\usepackage[T1]{fontenc}    %
\usepackage{hyperref}       %
\usepackage{url}            %
\usepackage{booktabs}       %
\usepackage{amsfonts}       %
\usepackage{nicefrac}       %
\usepackage{microtype}      %
\usepackage{xcolor}         %
\usepackage{booktabs}
\usepackage{multirow}
\usepackage{algorithm}
\usepackage{algpseudocode}
\usepackage[labelfont=bf,font=small]{caption}
\usepackage{wrapfig}
\usepackage{graphicx}
\usepackage{enumitem}
\usepackage{fontawesome}

\usepackage{basecommon}

\usepackage{inconsolata}

\title{Fast Matrix Multiplications for  Lookup Table-Quantized LLMs}

\author{
$\textbf{Han Guo}^{\star}$
\hspace{10.0mm} $\textbf{William Brandon}^\star$
\hspace{10.0mm} $\textbf{Radostin Cholakov}^{\dagger}$
\vspace{2mm} \\
$\textbf{Jonathan Ragan-Kelley}^{\star}$
\hspace{10.0mm} $\textbf{Eric P. Xing}^{\diamond}$
\hspace{10.0mm} $\textbf{Yoon Kim}^{\star}$ \vspace{2mm} \\
\normalfont
\textsuperscript{$\star$}Massachusetts Institute of Technology, \textsuperscript{$\dagger$}High School of Mathematics Plovdiv \\
\textsuperscript{$\diamond$}Carnegie Mellon University, MBZUAI, Petuum Inc. \\
\small{\texttt{\{hanguo,wbrandon,radi\_cho,jrk,yoonkim\}@mit.edu}}, \,\, \small{\texttt{epxing@cs.cmu.edu}} \vspace{2mm} \\
   \faGithub \,\,\,  {  \url{https://github.com/HanGuo97/flute}}
}

\begin{document}
\maketitle
\begin{abstract}
\vspace{-1mm}
The deployment of large language models (LLMs) is often constrained by memory bandwidth, where the primary bottleneck is the cost of transferring model parameters from the GPU's global memory to its registers. When coupled with custom kernels that fuse the dequantization and matmul operations, weight-only quantization  can thus enable faster inference by reducing the amount of memory movement. However, developing high-performance kernels for weight-quantized LLMs  presents substantial challenges, especially when the weights are compressed to non-evenly-divisible bit widths (e.g., 3 bits) with non-uniform, lookup table (LUT) quantization. This paper describes \texttt{FLUTE}, a \textbf{f}lexible \textbf{l}ook\textbf{u}p \textbf{t}able \textbf{e}ngine for LUT-quantized LLMs, which uses  offline restructuring of the quantized weight matrix to minimize bit manipulations associated with unpacking, and  vectorization and duplication of the lookup table to mitigate shared memory bandwidth constraints. At batch sizes < 32 and quantization group size of 128 (typical in LLM inference), the \texttt{FLUTE}  kernel can be 2-4$\times$ faster than existing  GEMM kernels.  As an application of \texttt{FLUTE}, we explore a simple extension to lookup table-based NormalFloat quantization  and apply it to quantize  LLaMA3 to various configurations, obtaining competitive quantization performance against strong baselines while obtaining an end-to-end throughput increase of  1.5 to 2 times.

\end{abstract}

\input{01-introduction}

\input{02-background}

\input{03-methods}

\input{04-experiments}

\bibliography{references}

\appendix

\input{appendix}

\end{document}

%% file: math-commands.tex
\usepackage{amsmath,amsfonts,bm}

\def\eqref#1{equation~\ref{#1}}

\def\1{\bm{1}}

\def\rmQ{{\mathbf{Q}}}

\def\rmS{{\mathbf{S}}}
\def\rmT{{\mathbf{T}}}

\def\rmW{{\mathbf{W}}}
\def\rmX{{\mathbf{X}}}
\def\rmY{{\mathbf{Y}}}

\DeclareMathAlphabet{\mathsfit}{\encodingdefault}{\sfdefault}{m}{sl}
\SetMathAlphabet{\mathsfit}{bold}{\encodingdefault}{\sfdefault}{bx}{n}

\DeclareMathOperator*{\argmin}{arg\,min}

\DeclareMathOperator{\clamp}{clamp}
\DeclareMathOperator{\round}{round}

%% file: 01-introduction.tex
\vspace{-4mm}
\section{Introduction}
\vspace{-2mm}

Large language model (LLM) deployment faces significant latency challenges due to the memory bandwidth constraints inherent in generative (token-by-token)  inference. The primary bottleneck is the cost of transferring model parameters from the GPU's global memory to the registers, i.e., LLM inference is \emph{memory-bound}. To overcome this ``memory wall'' \cite{gholami2024ai}, practitioners have increasingly adopted weight-only quantization methods, wherein  the parameters of an LLM are compressed to lower precision (e.g., 4 or 8 bits) than the precision in which they were trained (typically 16 bits). In addition to latency improvements, weight quantization can also drastically reduce GPU memory required for deployment.

Realizing practical speed-ups with weight-only quantization requires custom mixed-type matrix-matrix multiply (matmul) kernels  which must  (1) move  a layer's quantized weights  from GPU off-chip DRAM to  on-chip SRAM, (2) \emph{de}quantize the weights to floating-point (\texttt{FP}) format (on chip), (3) perform the \texttt{FP} matmul, and (4) write the results back to DRAM. Existing kernels such as \texttt{bitsandbytes} \cite{dettmers2024qlora}, \texttt{Marlin} \cite{frantar2024marlin}, and \texttt{BitBLAS} \cite{Microsoft} demonstrate that this strategy can result in significant matmul speed-ups, e.g. up to four times faster  when going from \texttt{W16A16} to \texttt{W4A16}. However, these kernels are typically specialized to 4-bit quantization, and while some kernels support non-uniform, lookup table (LUT) quantization, they  are generally slower than the uniform counterparts.
Given the recent promising results with odd-bit \cite{shao2023omniquant,ma2024affinequant,ma2024era} and non-uniform \cite{guo2023lq,kim2023squeezellm}  quantization methods, there is thus a need to develop  flexible kernels that can support mixed-type matmuls with a wider range of  settings.

This paper describes  \texttt{FLUTE}, a \textbf{f}lexible \textbf{l}ook\textbf{u}p-\textbf{t}able \textbf{e}ngine    for deploying weight-quantized LLMs, with a focus on the low-bit and non-uniform quantization setting. This setting raises several  challenges. First, going beyond 8-bit quantization  involves packing sub-8-bit matrices into supported data types, followed by unpacking during dequantization. Structuring the unpacked data to match GPU-native matmul formats is especially challenging when the weights are quantized to non-standard bit-widths. Second,  while uniformly-quantized models can rely on assembly-level optimizations to convert from \texttt{INT} to \texttt{FP} through bit-level manipulations,  lookup table-based dequantization involves dynamic indexing, and a na\"{i}ve implementation   can   lead to substantial overhead. Finally,  typical matmul implementations which distribute the workload across a grid of parallel thread blocks become inefficient with small batches and low bit-width weights; this  necessitates more sophisticated partitioning strategies to optimize hardware resource utilization.

\texttt{FLUTE} addresses these challenges through a combination of (1) offline weight restructuring, (2)  a shared-memory lookup table for efficient dequantization, and (3)  Stream-K partitioning  for optimized workload distribution. We compare \texttt{FLUTE} against existing  kernels on standard LLM mixed-precision matmul settings where weights are quantized to 4 bits in groups of 128, and find that it  outperforms existing non-uniform quantization kernels, and even matches the simpler uniform-quantization kernels in some cases. As an application of \texttt{FLUTE}, we experiment with quantizing LLaMA3---which has  been  found to be difficult to quantize \cite{huang2024good}---using a variant of normal float (\texttt{NF}) quantization \cite{dettmers2024qlora} which {learns} the quantization parameters based on calibration data. We find that we can achieve a 1.5 to 2 times increase in end-to-end throughput when integrated with  frameworks such as \texttt{vLLM} \cite{kwon2023efficient}.

%% file: 02-background.tex
\vspace{-2mm}
\section{Background and Related Work}
\vspace{-2mm}
\subsection{GPU Architecture and Memory Bandwidth Bottlenecks}
\vspace{-2mm}
GPUs are massively-parallel processors designed for throughput-oriented workloads containing large amounts of independent work. The hardware of a current-generation NVIDIA GPU consists of an array of many individual \emph{streaming multiprocessors} (``SMs''), each consisting of $4$ separate \emph{warp schedulers} together with a single \emph{shared memory}  scratchpad accessible to all $4$ warp schedulers. Each warp scheduler executes instructions on its own \emph{functional units}, and is able to issue at most one instruction per cycle, which may then take multiple subsequent cycles to complete while the warp scheduler moves on to concurrently issue other instructions. At the hardware level, the instructions executed by a warp scheduler typically operate in a SIMD fashion over vectors of $32$ data elements at a time.  At the software level, CUDA asks the programmer to program at the level of individual logical \emph{threads} executing scalar operations; threads are assigned sequential integer IDs, and every group of $32$ consecutive threads are implicitly organized together into a single \emph{warp}, corresponding to the GPU hardware's actual native unit of instruction execution.

Although GPUs are able to execute large numbers of instructions in parallel across the warp schedulers of their many SMs, the rate at which instructions can be executed is not always the bottleneck in realistic GPU workloads. Instead, the maximum achievable throughput of a GPU workload is often constrained by the speed of \emph{data movement} between levels of the GPU's memory hierarchy. The memory resources of modern NVIDIA server-class GPUs consist of (roughly): (1) Tens of gigabytes of off-chip DRAM, referred to here as \emph{global memory}; (2) Tens of megabytes of on-chip SRAM acting as a shared \emph{L2 cache} accessible to all SMs; (3) Hundreds of kilobytes of local SRAM per SM, split into two configurably-sized portions, one acting as an \emph{L1 cache} and the other an explicitly-addressed \emph{local scratchpad}; and (4) Hundreds of kilobytes of local SRAM per SM, acting as \emph{registers} for the threads running on that SM.

The read/write bandwidth of resources in this memory hierarchy can easily become the limiting factor for realistic GPU workloads. For example, an A100-80GB GPU supports a nominal peak throughput for 16-bit matrix-multiply instructions of $\approx 3 \times 10^{14}$ FLOP/s (aggregated across all SMs), but its main memory supports a nominal peak bandwidth of only $\approx 1.5 \times 10^{12}$ byte/s. This means that the speed of any kernel which performs fewer than roughly $(3 \times 10^{14}) / (1.5 \times 10^{12}) = 200$ matrix-multiply FLOPs per byte of data accessed will necessarily be limited by the GPU's memory bandwidth, not by its compute throughput. 
Maximizing the ratio of FLOPs to bytes transferred, a quantity known as \emph{arithmetic intensity}, is often the single most important consideration when designing high-performance kernels.

\vspace{-2mm}
\subsection{LLM Deployment Characteristics}
\vspace{-1mm}
Depending on the context, inference can be bottlenecked by compute throughput  or memory bandwidth.  For LLMs, training, large-prefill, and large-batch inference  enjoy high arithmetic intensity as  the sizes of matrices involved in the matmuls are  large enough to saturate compute. Small-batch, token-by-token inference on the other hand involves narrower matmuls due to the smaller batch dimension, resulting in low arithmetic intensity. Reducing the amount of memory operations in this case can thus enable practical speed-ups, even if the number of FLOPs remains the same (or is even slightly increased). 
This has led to much recent work on customized kernels which  move the  weights from main memory to on-chip SRAM while keeping them quantized/sparse \cite{dettmers2024qlora,kim2023squeezellm,frantar2024marlin,Microsoft,xia2024flash},  and then performing the actual matmuls in higher precision after dequantizing to \texttt{FP} on chip. \texttt{Marlin} implements this strategy for 4-bit uniform  quantization and reports significant (up to 4$\times$) matmul speed-ups   even in moderate batch (16-32) settings. \texttt{bitsandbytes} \cite{dettmers2024qlora} and \texttt{BitBLAS} \cite{Microsoft} extend this to LUT-quantized LLMs, but  do not allow for 3 bit-quantized weights. Moreover, existing LUT-quantization kernels generally underperform uniform-quantization kernels.

\vspace{-2mm}
\subsection{Weight-only Quantization in LLMs}
\vspace{-1mm}
Uniform quantization converts a group of full precision weights to lower-precision intervals of equal size through rounding. For example min-max  quantization maps a group of weights $\boldu$ to integers $\{-2^{b-1}, \dots, 2^{b-1} - 1 \}$ via the function  $\clamp\left(\round(\frac{1}{s}  \boldu ) ; -2^{b-1}, 2^{b-1} - 1\right)$,
where $s = \frac{\max(|\boldu|)}{2^{b-1}-1}$ is a scaling factor.    Recent methods improve upon min-max quantization by using calibration data \cite{frantar-gptq,lin2023awq,shao2023omniquant,ma2024affinequant}.
When both the weights and activations are quantized uniformly, it is possible to use  \texttt{INT} matmuls to enable speed-ups beyond the savings from reduced memory movement.
However, activation quantization remains difficult due to the presence of outlier channels, which necessitate sophisticated mitigation strategies \cite{wei2022outlier,dettmers2022gpt3,xiao2022smoothquant,zhao2023atom,ashkboos2023towards,ashkboos2024quarot,nrusimha2024mitigating,lin2024qserve}.  Weight-only quantization thus remains a popular choice for LLMs. Moreover, if only the weights are quantized, it is possible to reduce quantization error further by  applying quantization at a more fine-grained levels (e.g., a block of 128 weight values) than at row- or column-level.

Non-uniform quantization generalizes uniform quantization by  mapping weights to potentially \emph{unequal} intervals \cite{miyashita2016convolutional,zhou2017incremental,zhang2018lq,yang2019quantization}.
Lookup table (LUT) quantization is a  flexible variant of non-uniform quantization which can map intervals to arbitrary values via a lookup table \cite{cardinaux2020iteratively,wang2022learnable}. LUT quantization needs to trade off the size of the lookup table and the granularity of the groups at which the weights are quantized. For example, SqueezeLLM \cite{kim2023squeezellm} applies  K-means clustering at the column (output channel) level to obtain the lookup table, while NormalFloat quantization \cite{dettmers2024qlora} uses a tensor-level lookup table obtained from the quantiles of a Normal distribution that is multiplicatively modified through  group-level parameters. While it is possible  to perform matmuls with activations/weights that are quantized non-uniformly (e.g., through LUT-based matmuls \cite{xu2021multiplication,park2022lut}), these methods cannot leverage specialized accelerators  on modern GPUs which are typically optimized for \texttt{FP} matmuls. 
We thus seek efficient kernels which can simultaneously make use of quantized representations (to minimize memory movement) as well as  GPU-native matrix multiplications in \texttt{FP}.

%% file: 03-methods.tex
\vspace{-1mm}
\section{\texttt{\mdseries FLUTE}: A Fast and Flexible Kernel for Mixed-Type Matrix Multiplications}
\vspace{-1mm}
Let   $\rmQ \in \mathbb{Z}^{k \times n}$ be a quantized matrix obtained from  quantizing the weight matrix $\rmW \in \mathbb{R}^{k \times n}$ using a lookup table $\mathbf{T}$. Concretely, given a lookup table  $ \mathbf{T} =  [v_0, \dots, v_{2^b-1}]$ where $b$ is the number of bits and each $v_i$ is a floating-point number,  each entry of $\rmQ$ is given by,
\begin{align*}
    \rmQ_{ij} = \operatorname{quantize}(\mathbf{W}_{ij}; \mathbf{T}) = \argmin_{c} |\mathbf{W}_{ij} - v_c|, \vspace{-1mm}
\end{align*}
where $\rmQ_{ij} \in \{0, \dots, 2^{b}{-1}\}$. Now let $\widehat{\rmW} \in \mathbb{R}^{k \times n}$ be the \emph{de}quantized matrix where
\begin{align*}
   \widehat{\rmW}_{ij} = \operatorname{dequantize}(\rmQ_{ij}, \mathbf{T}) = \mathbf{T}[\rmQ_{ij}].
\end{align*}
Our objective is to perform a fast matrix multiplication between a dense input activation matrix $\rmX \in \mathbb{R}^{m \times k}$ (typically stored in \texttt{FP16}) and $\widehat{\rmW}$.

\input{tables/algorithm-simplified}

A  straightforward implementation of such mixed-type matrix  multiplication uses separate kernels. The first kernel loads the quantized matrix $\rmQ$ from the GPU's off-chip global memory into its on-chip memory, performs dequantization, and writes back the dequantized matrix $\widehat{\rmW}$ back to the DRAM. The second kernel  is a standard \texttt{FP} matmul kernel over $\rmX$ and $\widehat{\rmW}$. This separate-kernel  can introduce substantial overhead since  $\widehat{\rmW}$ is moved back and forth. We can achieve faster matmuls by \emph{fusing} the dequantization and matmul kernels, where we dequantize on chip and immediately use the dequantized values for the matmul. 

However, implementing a fused weight-only LUT-quantized matmul that leads to  speed-ups presents several challenges. For one, high-performance matmul necessitates the use of specialized primitives, such as Tensor Cores, which have strict requirements regarding the types, shapes, and layout of data. Second, efficient dynamic indexing is crucial for LUT-based dequantization; however, GPUs do not natively support dynamic indexing of a lookup table in their fastest on-chip registers. Finally, with smaller input matrices arising from low-bit and low-batch deployment,  achieving workload balance across SMs is vital for maintaining speed, thus necessitating sophisticated partitioning strategies. \texttt{FLUTE} addresses these challenges through a combination of offline restructuring of the quantized weight matrix (\S\ref{ssec:offline}), vectorization and duplication of the lookup table to mitigate shared bandwidth constraints (\S\ref{ssec:bandwidth}), and Stream-K workload partitioning to minimize wave quantization (\S\ref{ssec:streamk}). Alg.~\ref{alg:flute-simple} gives a simplified version of the \texttt{FLUTE} kernel, while Fig.~\ref{fig:basic-kernel-design} shows a high-level overview. (See Alg.~\ref{alg:flute} in the Appendix for more details).

\begin{figure}[t]
\centering
\includegraphics[width=.99\linewidth]{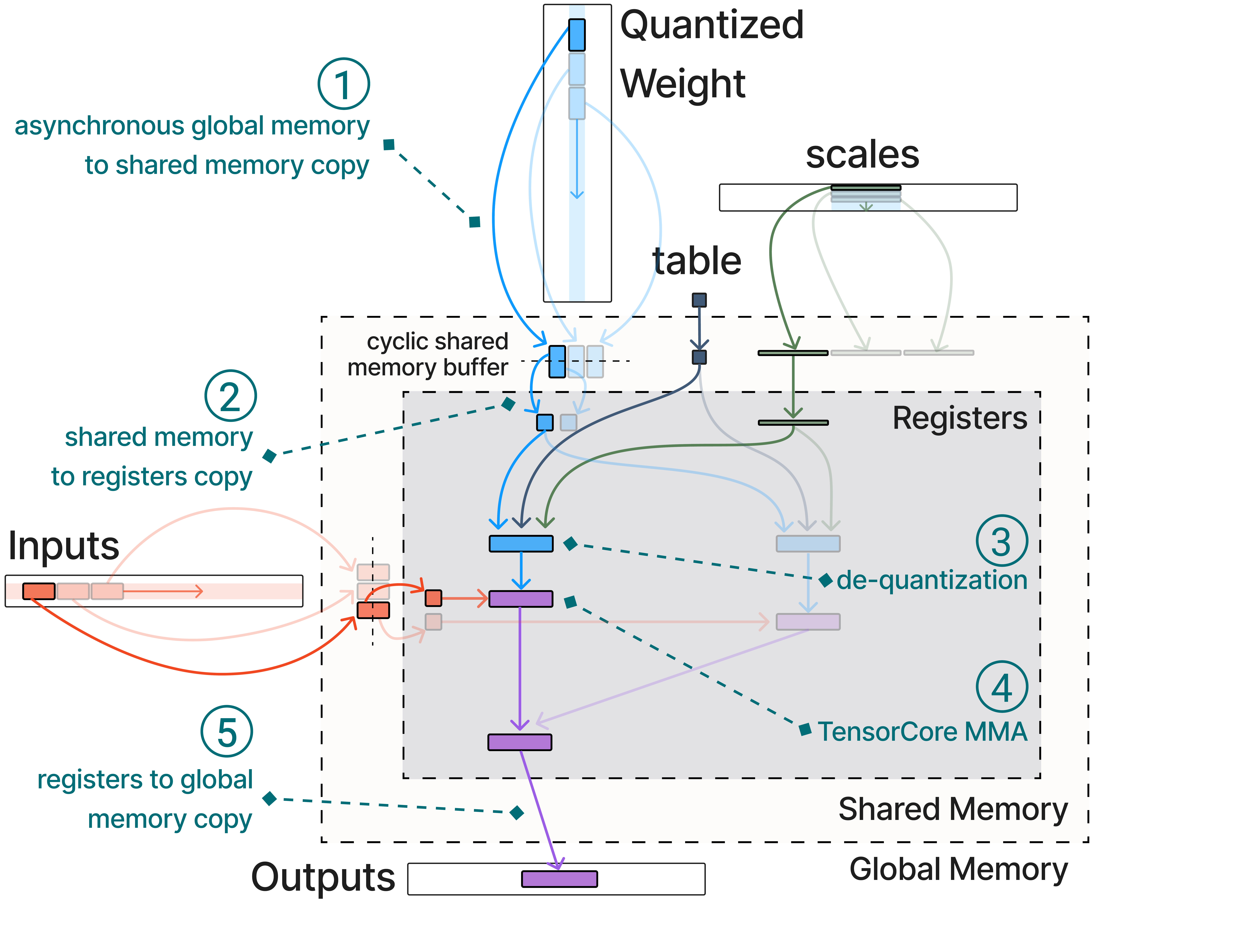}
\vspace{-4mm}
\caption{A simplified view of a kernel that fuses the dequantization and matmul steps. Each threadblock (group of threads) is responsible for computing one or more output tiles by performing the matrix product between specific rows of inputs and columns of weights. {\color[HTML]{006d77} \textbf{(1)}} The threadblock issues asynchronous copy instructions to fetch small chunks of input data (tiles) from global memory to shared memory. {\color[HTML]{006d77} \textbf{(2)}} As soon as a tile arrives in shared memory, it is further sliced into smaller chunks (fragments) and copied into registers. {\color[HTML]{006d77} \textbf{(3)}} Once all necessary components are in the registers, the quantized matrix undergoes dequantization. {\color[HTML]{006d77} \textbf{(4)}} The dequantized matrix and inputs are then processed by Tensor Cores using MMA (Matrix Multiply Accumulate) instructions. {\color[HTML]{006d77} \textbf{(5)}} Finally, the accumulated results are written back from the registers to the outputs in global memory.
}
\label{fig:basic-kernel-design}
\vspace{-3.6mm}
\end{figure}

\begin{figure*}[t]
\centering
\begin{minipage}[b]{0.3\textwidth}
\includegraphics[width=\textwidth]{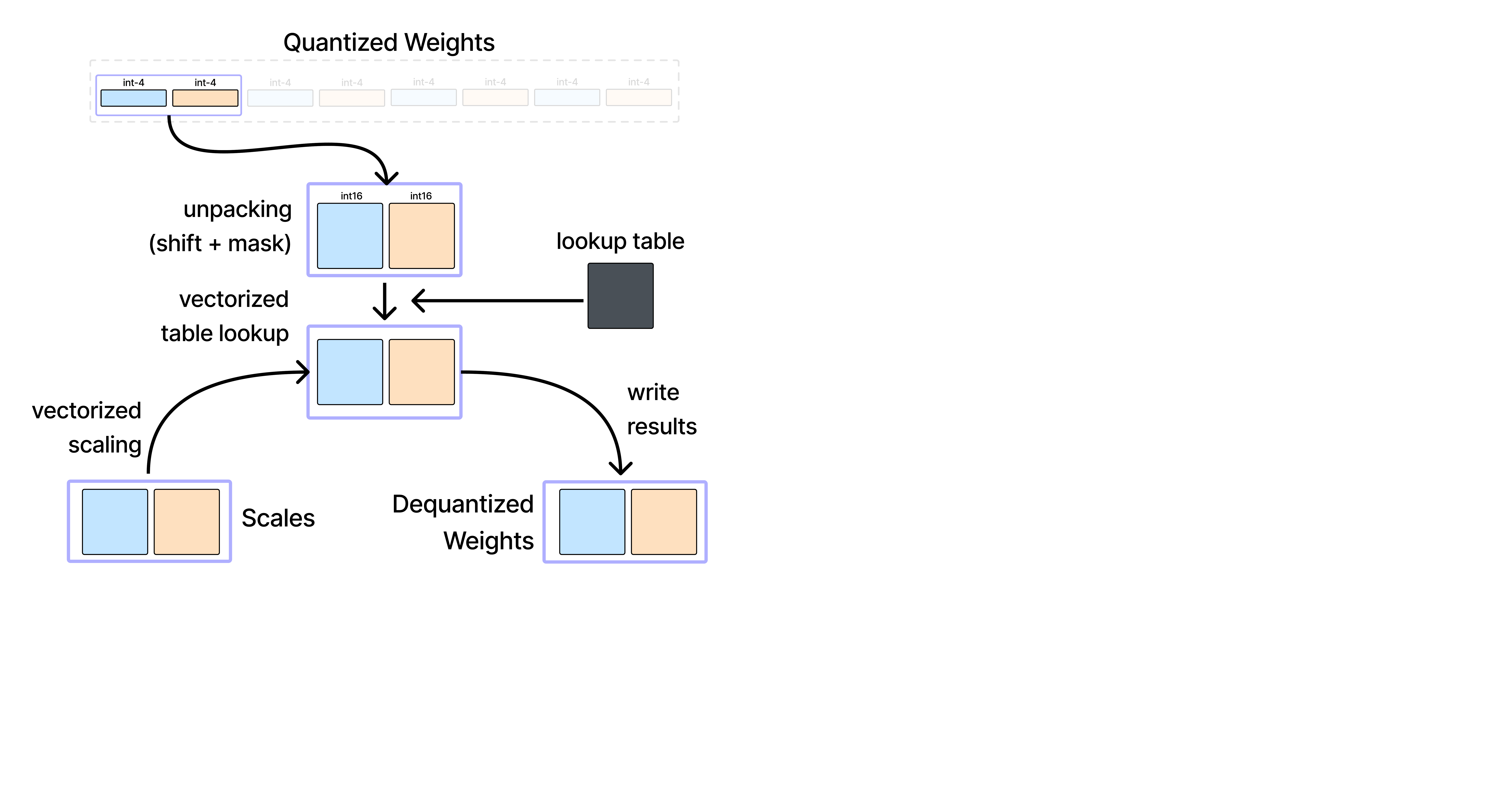}
\end{minipage}
\hfill
\begin{minipage}[b]{0.69\textwidth}
\includegraphics[width=\textwidth]{figs/stream-k-small.pdf}
\end{minipage}
\caption{\textbf{Vectorized Lookup Table Design (Left).} Instead of dequantizing one element at a time, we vectorize the lookup table by creating another table that holds the values of all possible pairs of indices. This can look up two values simultaneously, followed by efficient vectorized scaling operations.
\textbf{Stream-K Work Decomposition (Right).} In classic work decomposition, output tile production is independently assigned to threadblocks. Each threadblock processes one (or more) rows of the left operand and one (or more) columns of the right operand, slicing down the inner K dimension to compute the corresponding output tile (Slice-K).  However, when the weight matrix is heavily quantized, the reduced size can lead to ``stragglers'' in Slice-K due to uneven workload assignment. Stream-K~\cite{osama2023stream} addresses this by decomposing work at a finer granularity, enabling multiple threadblocks to collaboratively compute a single output tile.}
\label{fig:vectorized-lut}
\label{fig:slice-k-and-stream-k}
\vspace{-4mm}
\end{figure*}

\vspace{-1mm}
\subsection{Offline Matrix Restructuring}
\vspace{-1mm}
\label{ssec:offline}
Modern GPUs feature specialized primitives (Tensor Cores)---distinct from general-purpose vector ALUs---which can substantially accelerate dense matrix multiplications. For example, A100's \texttt{FP16} tensor core matmuls are  16$\times$ faster than \texttt{FP32} vector matmuls. However, this acceleration comes at the expense of generality and programmability. Tensor Core \texttt{MMA} (matrix-multiply-accumulate) operations require the input matrices to adhere to specific layout specifications within the registers of 32 threads. The fused kernel needs to first load fragments of the quantized weight into registers, dequantize the matrix, and then perform the \texttt{MMA} operation between the input fragments and dequantized matrix fragments. This necessitates that the {post-dequantization} matrix layout  meets the required specifications. While runtime data reordering is one approach, it introduces a substantial number of operations. Instead, we leverage the fact that $\rmQ$ (the quantized weights) are static during inference, allowing for offline weight reordering such that after dequantization, the weights are already laid out exactly in the expected format~\cite{frantar2024marlin,xia2024fp6,lin2024qserve}.

The above strategy is difficult to straightforwardly extend to the case of non-evenly-divisible bit widths (e.g., 3 bits). Kernels  employ vectorized data access when loading data from global  to shared memory. Hence each thread should access the quantized weight in granularity of 128 bits (or at least in powers of 2). While this could be addressed by padding, this would be inefficient. We instead split the (3-bit) quantized weight into two partitions~\cite{xia2024fp6}, or bit-slices: one containing the 1-bit portion and the other the 2-bit portion, and issue two separate vectorized (asynchronous) data copy instructions. Once the two bit-slices are loaded into the registers, we  combine them before  dequantization.

\vspace{-2mm}
\subsection{Vectorized Lookup in Shared Memory}
\vspace{-1mm}
\label{ssec:bandwidth}
During dequantiation each element $c$ in the quantized array needs to access a $16$-bit  element $\mathbf{T}[c]$ from the lookup table. Each thread  needs to access the lookup table using different indices, and such ``non-uniform access''  can  degrade runtime performance when implemented na\"{i}vely.  While storing the lookup table in on-chip shared memory can reduce expensive off-chip memory access, this still introduces significant traffic to shared memory.  To reduce memory access instructions we ``vectorize'' the lookup operation by accessing two elements at a time (Fig.~\ref{fig:vectorized-lut}, left). We build an alternative lookup table for every possible \emph{pair} of values, with each element containing a tuple of 16-bit values. The storage overhead from the vectorized lookup table is minimal compared to the rest of memory usage. For example, for 4-bit LUT the table has only $2^4$ elements of $16$-bit values, and thus the vectorized table has only $2^8$ elements of $32$-bit values, slightly more than 1KB of storage. This is a  fraction of the 48KB-163KB of shared memory on modern GPUs.

\vspace{-2mm}
\paragraph{Reducing bank conflicts.} Shared memory is organized such that each successive 32-bit segment corresponds to a ``bank'', and there are 32 such banks. Each memory address in shared memory corresponds to the $\lfloor \frac{\operatorname{addr}}{32}\rfloor \operatorname{mod} 32$ bank. If threads in a warp access data from different banks,  access is parallelized. However, if two threads access data from the same bank (but not the same address), the access is serialized. For the 4-bit and 3-bit vectorized tables, a simple implementation could thus cause up to 8-way bank conflicts (4-bit) or 2-way bank conflicts (3-bit). To mitigate this, we  duplicate the 4-bit vectorized lookup table multiple times, placing copies in different memory banks, which allows threads to access values from different banks with reduced bank conflicts.

\begin{figure*}[t]
\centering
\includegraphics[width=.99\textwidth]{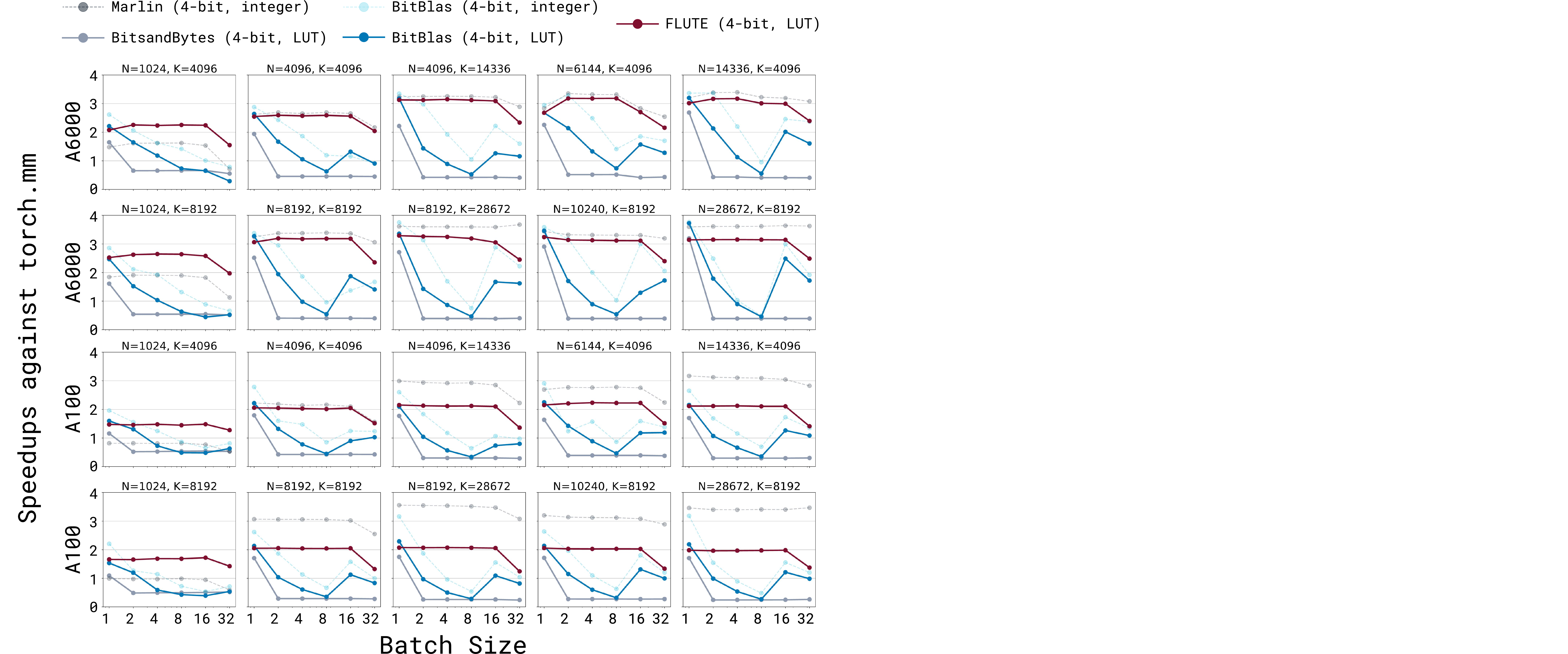}
\caption{Runtime performance of \texttt{FLUTE} in the standard W4G128 setting where, the weights are quantized to 4 bits in groups of 128. We show speedup against 16-bit \texttt{torch.mm}.  The matrix shapes for our benchmarks are selected based on those used in Llama-3-8B (top row) and Llama-3-70B (bottom row) models. For each M-N-K shape tuple, we generate three random sets of data, run each kernel on the data 100 times, and average. While our main comparisons are against other LUT kernels (\texttt{bitsandbytes}, \texttt{BitBLAS-NF4}), for reference we also include comparisons with kernels that only support uniform (integer) dequantization (\texttt{Marlin}, \texttt{BitBLAS}). These results are represented with dashed lines in our figures. 
}
\label{fig:microbenchmarks}
 \vspace{-5mm}
\end{figure*}

\vspace{-2mm}
\subsection{Stream-K Workload Partitioning}
\vspace{-1mm}

\label{ssec:streamk}
For high SM occupancy, standard matmul implementations block the computation using a data-parallel tiling of the output matrix, where a group of threads ( ``thread block'') is assigned to compute the work on one output tile. This is shown on the left of Figure~\ref{fig:slice-k-and-stream-k}. As each thread block can only occupy one SM, it is important to avoid ``wave quantization'', which happens when the number of output tiles is not an even multiple of the number of processor cores. In this case the last wave uses only a subset of the cores, leaving the rest idle.

Wave quantization and workload imbalance are especially problematic in low-bit and low-batch scenarios, {which result in smaller input matrices (activations and quantized weights)}, thus making the effect of wave quantization more pronounced.
To mitigate this, we implement a method known as Stream-K workload decomposition~\cite{osama2023stream}, which distributes the tiles such that each SM's computations can span beyond specific rows or columns. This method is depicted on in Fig.~\ref{fig:slice-k-and-stream-k} (Right). Here, the 35 M-N-K tiles are more evenly divided among the 3 SMs than in the simpler Slice-K partitioning (Figure~\ref{fig:slice-k-and-stream-k}, middle), in which SM's computations do not span beyond rows/columns.

\vspace{-2mm}
\paragraph{Mixed precision accumulation and global reduction.}
In Stream-K, when multiple SMs compute the same M-N dimension across different K tiles, they must reconcile their partial sums in  off-chip global memory. SMs that complete their share of K tiles write their partial sums to a global scratch space, allowing subsequent SMs to read, reduce, and write back these sums.
For numerical stability, most kernels perform multiplications in \texttt{FP16} but accumulate results in \texttt{FP32}. However, writing to global memory in \texttt{FP32} results in significant traffic. We thus implement in-register accumulation in \texttt{FP32} and globally reduce partial sums in \texttt{FP16}.

%% file: tables/algorithm-simplified.tex
\algblock{ParFor}{EndParFor}
\algnewcommand\algorithmicparfor{\textbf{parallel for}}
\algnewcommand\algorithmicpardo{\textbf{do}}
\algnewcommand\algorithmicendparfor{\textbf{end\ parallel for}}
\algrenewtext{ParFor}[1]{\algorithmicparfor\ #1\ \algorithmicpardo}
\algrenewtext{EndParFor}{\algorithmicendparfor}

\begin{figure}[t]
 \vspace{-4mm}
\begin{minipage}{0.99\linewidth}
\begin{algorithm}[H]
\ttfamily
\scriptsize
\captionsetup{font=footnotesize}
\caption{$\texttt{FLUTE (Simplified)}$}
\label{alg:flute-simple}
\begin{algorithmic}
\Require
$\rmX^{\text{g}}$: inputs in HBM \\
\quad\quad\quad\
$\rmQ^{\text{g}}$: quantized weight in HBM \\
\quad\quad\quad\,
$\rmS^{\text{g}}$: scales in HBM \\
\quad\quad\quad\,
$\rmT^{\text{g}}$: lookup table in HBM \\
\quad\quad\quad\,
$\rmY^{\text{g}}$: outputs in HBM \\
\State {\color{gray} \# --------------- Terminology and Shapes (of $\rmX$ only for brevity) ---------------}
\State {\color{gray} \# tile: a block of matrix entries that fits into shared memory}
\State {\color{gray} \# fragment: a block of tile entries that fit into registers}
\State {\color{gray} \# g(lobal), s(shared), r(egisters) denotes where data reside}
\State {\color{gray} \# $\rmX^\text{g}$: [M tiles, K tiles, fragments per tile, fragment size]}
\State {\color{gray} \# $\rmX^\text{s}$: [fragments per tile, fragment size]}
\State {\color{gray} \# $\rmX^\text{r}$: [fragment size]}
\State {\color{gray} \# --------------- Offline Preprocessing and Host-side Code ---------------}
\State $\rmQ^\text{g}_1, \rmQ^\text{g}_2  \gets \texttt{reorder\_and\_split}(\rmQ^\text{g})$ \Comment{Section~\ref{ssec:offline}}
\State $\rmT^\text{g}_\text{v} \gets \texttt{make\_vectorized\_LUT}(\rmT^\text{g})$ \Comment{Section~\ref{ssec:bandwidth}}
\State $\texttt{tile\_scheduler} \gets \texttt{StreamKTileScheduler}(\text{N}_{\text{SMs}})$ \Comment{Section~\ref{ssec:streamk}}
\State {\color{gray} \# launching blocks proportional to SMs}
\State $\text{N}_{\text{blocks}} \gets \texttt{tile\_scheduler.num\_blocks()}$
\State {\color{gray} \# --------------- Kernel Launches and Device-side Code ---------------}
\ParFor{block index b $\gets$ 1 to $\text{N}_{\texttt{blocks}}$} 
    \State {\color{gray} \# partitioning works for this block}
    \State \texttt{tile\_scheduler.initialize(b)}
    \State {\color{gray} \# copy lookup table from global to shared memory}
    \State $\texttt{copy}^{\texttt{g}\rightarrow\texttt{s}}(\rmT^\text{g}_\text{v}, \rmT_\text{v}^{s})$
    \State {\color{gray} \# initialize the register-backed accumulator}
    \State $\rmY^\text{r} \gets \mathbf{0}$
    \State {\color{gray} \# main loop (pipelined in practice, not shown here)}
    \While{$\neg\texttt{tile\_scheduler.done}()$}
        \State {\color{gray} \# copy data tile  from global to shared memory}
        \State {\color{gray}\#  ($\rmX$ is shown but the same applies to $\rmQ_1, \rmQ_2, \rmS$)}
        \State $\text{i}_\text{tile},\text{j}_\text{tile},\text{k}_\text{tile} \gets \texttt{tile\_scheduler.get\_tile\_index}()$
        \State $\texttt{copy}^{\texttt{g}\rightarrow\texttt{s}}(\rmX^\text{g} [\text{i}_\text{tile}, \text{k}_\text{tile}, \text{:}, \text{:}], \rmX^{\text{s}})$
        \For{fragment index $\text{t}_\text{fragment} \gets $ 1 to $\text{N}_{\text{fragments}}$}
            \State {\color{gray} \# copy data fragment from shared memory to registers}
            \State {\color{gray}\#   ($\rmX$ is shown but the same applies to $\rmQ_1, \rmQ_2, \rmS$)}
            \State $\texttt{copy}^{\texttt{s}\rightarrow\texttt{r}}(\rmX^\text{s} [\text{t}_\text{fragment}, \text{:}], \rmX^\text{r})$
            \State {\color{gray} \# combine two bit-slices in registers (3-bit)}
            \State $\rmQ^\text{r}_\text{1+2} \gets \text{combine}(\rmQ^\text{r}_\text{1}, \rmQ^\text{r}_\text{2})$ \Comment{Section~\ref{ssec:offline}}
            \State {\color{gray} \# vectorized dequantization in registers}
            \State $\widehat{\rmW}^\text{r} \gets \text{vec\_dequantize}(\rmQ^\text{r}_\text{1+2}, \rmS^\text{r}, \rmT^\text{s})$ \Comment{Section~\ref{ssec:bandwidth}}
            \State {\color{gray} \# i.e., $\rmY^{\text{r}} \gets \rmY^{\text{r}} + \rmX^{\text{r}} \widehat{\rmW}^{\text{r}}$}
            \State $\rmY^{\text{r}} \gets \text{tensor\_core\_mma}(\rmY^{\text{r}}, \rmX^{\text{r}}, \widehat{\rmW}^{\text{r}})$
        \EndFor
        \If{tile\_scheduler.end\_of\_output\_tile()}
            \State $\widehat{\rmY}^\text{r} \gets \text{to\_fp16}(\rmY^\text{r})$ \Comment{Section~\ref{ssec:streamk}}
            
            \State {\color{gray} \# write output tile from Registers to HBM}
            \State $\text{copy}^{\text{r}\rightarrow\text{g}}(\widehat{\rmY}^\text{r}, \rmY^\text{g}[\text{i}_\text{tile}, \text{j}_\text{tile}, \text{:}])$
        \EndIf
        \State {\color{gray} \# update the internal counter of scheduler}
        \State $\texttt{tile\_scheduler.step}()$
    \EndWhile
\EndParFor
\end{algorithmic}
\end{algorithm}%
\end{minipage}
\vspace{-6mm}
\end{figure}

%% file: 04-experiments.tex
\vspace{-2mm}
\section{Experiments}
\vspace{-2mm}

\begin{figure*}[t]
\centering
\includegraphics[width=.99\textwidth]{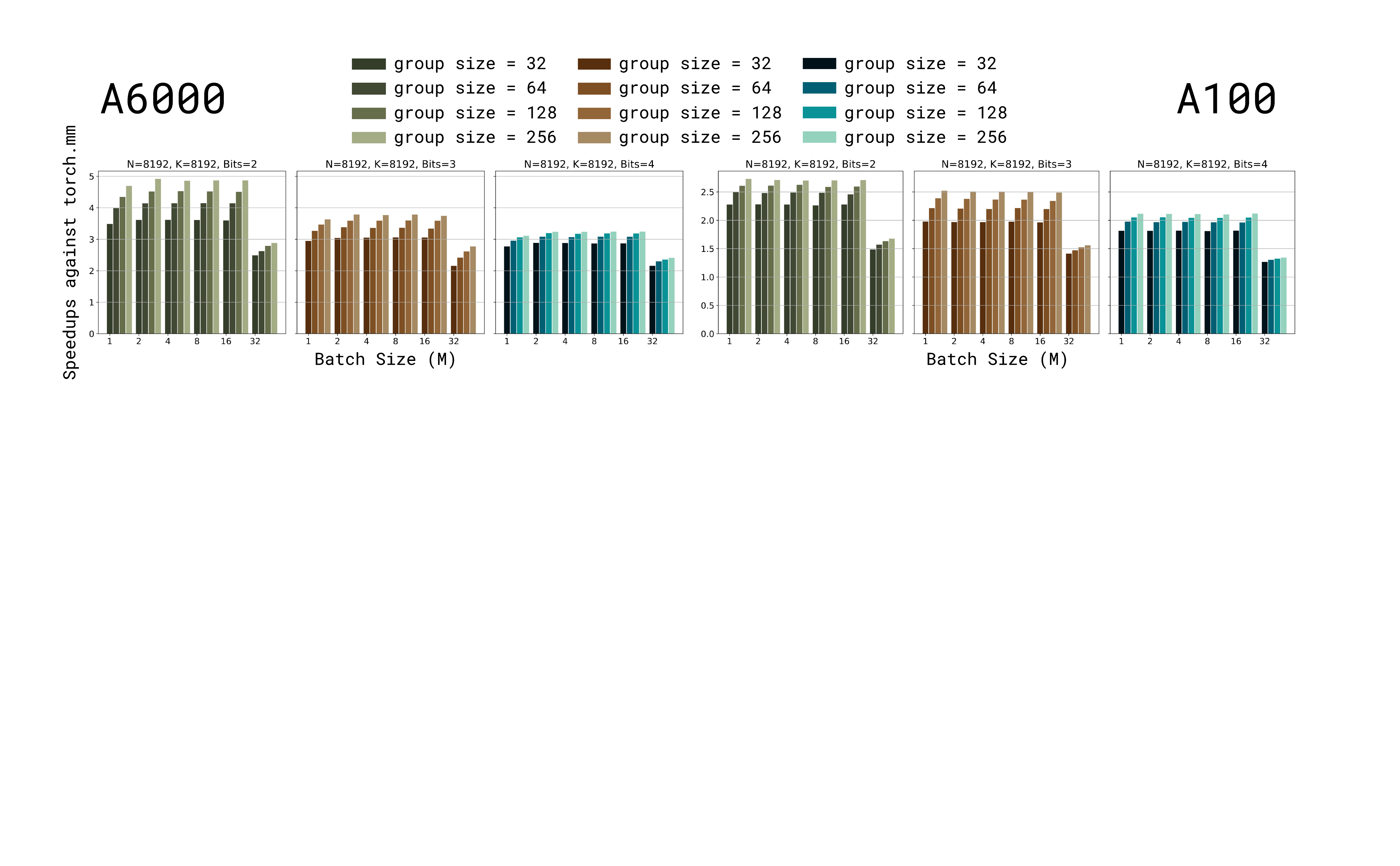}
\caption{Runtime performance at various bit-widths and group sizes with \texttt{N=K=8192}. \texttt{FLUTE} consistently achieves speedups across different settings, including the in the 3-bit configuration.}
\label{fig:microbenchmarks-bits-and-groupsizes}
\vspace{-4mm}
\end{figure*}

Our experiments consist of two settings:  \emph{kernel-level} experiments which compare \texttt{FLUTE} matmuls standalone against existing mixed-input matmul kernels (\S\ref{sec:kernel-exp}), and \emph{end-to-end} experiments  which assess whether practicals speed-ups are obtainable on realistic LLM workloads  (\S\ref{sec:e2e-exp}). 

\vspace{-2mm}
\subsection{Kernel Benchmarks}
\vspace{-1mm}
\label{sec:kernel-exp}

For each matrix size, we compile multiple instantiations of the kernel with various configurations, including different tile sizes, pipeline stages, and the number of lookup table duplicates, selecting the best-performing configuration based on benchmarking.\footnote{Concretely, we randomly generate three sets of input data based on the matrix shapes for 8B/70B LLMs, run the kernel on the data 100 times, and report the average performance on both A100 and A6000 GPUs.} We compare \texttt{FLUTE} against a collection of weight-quantized matrix multiplication kernels, including those capable of flexible LUT-based dequantization such as \texttt{bitsandbytes}~\cite{dettmers2024qlora}\footnote{For most of the kernels, we pre-allocate the output memory buffer and use the \texttt{out} keyword to exclude the memory allocation time from our measurements. However, as of this writing, \texttt{bitsandbytes} still allocates memory in some cases. Our preliminary experiments indicate that this introduces an overhead of approximately $2.5\%$.} and \texttt{BitBLAS}~\cite{Microsoft}. 

\vspace{-2mm}
\paragraph{LUT quantization method.} There are many methods for LUT quantization; we follow the popular NormalFloat LUT quantization scheme \cite{dettmers2024qlora}, where a tensor-level  table $\mathbf{T}$ is modified to be a group-level  table via a scaling factor $s_g$ for each group $g$, resulting in a group-level table $\mathbf{T}_g = [s_g \cdot v_0, \dots, s_g  \cdot v_{2^b-1}]$. Here $s_g \in \reals^{+}$ is a scalar that varies per group. This approach requires maintaining a tensor-level lookup table $\mathbf{T}$ and group-level scalars $\{s_{g}\}_{g=1}^{d^2 / g}$, and thus incur almost the same memory overhead as uniform quantization (which also requires maintaining the group-level scalars).
While we primarily focus on LUT kernels, for completeness we also compare against high-performance kernels specialized for uniformly quantized weights (\texttt{BitBLAS}\footnote{\url{https://github.com/microsoft/BitBLAS}} and \texttt{Marlin}\footnote{\url{https://github.com/IST-DASLab/marlin}}~\cite{frantar2024marlin}). These kernels do not require dynamic indexing into a lookup table and can perform dequantization in registers using highly tuned PTX assembly instructions that are not applicable to LUT-based dequantization.

\vspace{-2mm}
\paragraph{Results.} Figure~\ref{fig:microbenchmarks} presents the results with the standard setting of 4-bit quantization and a group size of 128, where memory traffic is reduced by 4\texttt{x} (modulo the overhead coming from scales). \texttt{FLUTE} achieves favorable performance across a wide range of matrix shapes on both A6000 and A100,  occasionally nearing the peak theoretical speedup (of 4\texttt{x}) on A6000. Other LUT-compatible kernels  achieve similar speedups only with a batch size of $1$, and their performance quickly degrades.  \texttt{FLUTE} also compares favorably to \texttt{Marlin}, which is highly specialized for cases where the input is \texttt{FP16} and the weight is uniform-quantized to \texttt{INT4}.

We further showcase the flexibility of \texttt{FLUTE} by experimenting with different group sizes not just in terms of its lookup-table design but also in supporting various bit-widths and group sizes. In particular, \texttt{FLUTE} can perform multiplications with 3-bit matrices (\S\ref{ssec:offline}), a capability that the aforementioned alternatives do not support. The results in Figure~\ref{fig:microbenchmarks-bits-and-groupsizes} demonstrate consistent speed-ups over \texttt{torch.mm} across across a wide rage of settings.

\vspace{-2mm}
\subsection{End-to-End LLM Benchmarks}
\vspace{-1mm}
\label{sec:e2e-exp}

As an application of \texttt{FLUTE}, we experiment with quantizing LLaMA3-8B and LLaMA3-70B. The LLaMA3 family of models has been found to be more difficult to quantize than other open source models \cite{chen2024efficientqat}, and thus presents a testing ground for different quantization strategies. 

For the LUT quantization method, we use a simple extension of {NormalFloat} (NF) quantization \cite{dettmers2024qlora}. Standard NF quantization calculates $2^{b-1}$ evenly-spaced  values from $[\delta, \frac{1}{2}]$, and $2^{b-1}+1$ evenly-spaced values from $[\frac{1}{2}, 1-\delta]$, where $\delta = \frac{1}{2}(\frac{1}{30} + \frac{1}{32})$. This results in $2^{b}$ probability values $[p_0, \dots, p_{2^{b}-1}]$ where $p_0 = \delta,  p_{2^{b-1}-1} = \frac{1}{2}$, and $p_{2^{b}-1} = 1-\delta$. These probabilities are converted into quantiles $[q_0, \dots, q_{2^{b}-1}]$ where $q_i = \Phi^{-1}(p_i)$ is the Gaussian quantile for $p_i$. The quantiles are then normalized to $[-1,1]$ by $\tilde{q}_{i} = \frac{q_i}{q_{2^{b}-1}}$.  Then, given a group of weights $\boldu=[u_1, \dots, u_B]$ and the absmax value $s = \max(|\boldu|)$ for that group, the weights $u_j$ in this group are quantized to the nearest quantile, i.e., 
    $c_{j} = \argmin_{i \in \{0, \dots, 2^{b}-1\}} \left|\tilde{q_i} - \frac{u_j}{s} \right|$. 
Given an NF-quantized matrix $\mathbf{Q} \in \{0, \dots, 2^b - 1\}^{k \times n}$, the matmul kernel loads the tensor-level lookup table $\mathbf{T} = [\tilde{q}_0, \dots \tilde{q}_{2^b-1}]$, as well as the group-level scales $s_1, \dots s_{\frac{kn}{B}}$, and then dequantizes via $\mathbf{T}[\mathbf{Q}_{ij}] \cdot s_{(i\times j) \mod B}$.

Our simple extension builds upon the above by using calibration data to refine the scales, which has been found to be beneficial for uniform quantization \cite{shao2023omniquant}.
Since the lookup table consists of quantiles from $\mathcal{N}\left(0,\sigma^2\right)$ with standard deviation $\sigma = \frac{1}{\Phi^{-1}(1 - \delta)}$, we can  reformulate the  quantization function as $c_{j} = \argmin_{i \in \{0, \dots, 2^{b}-1\}} \left|s\tilde{\sigma}q_i -u_j \right|$. For learning, we initialize    $\tilde{\sigma}=\frac{1}{\Phi^{-1}(1 - \delta)}$ and optimize this with gradient descent against the negative log-likelihood of calibration samples, where we use the straight-through estimator.\footnote{We also experimented with a variant of this approach where each value of the tensor-level lookup table is updated to be the average of all the weights that were bucketed to that value (as in K-means). We did not find meaningful improvements with this approach.} After learning, we can save  $\frac{s\tilde{\sigma}}{\sigma}$ as the new scale, and hence the number of scalar values to be loaded for dequantization remains unchanged. We use use 128 examples of  length 2048  from  WikiText-2 training  as our calibration dataset.

\input{tables/group-sizes}

We conducted end-to-end evaluations by integrating the \texttt{FLUTE} kernels into two libraries:
\begin{enumerate}
    \item \texttt{GPT-Fast}\footnote{\url{https://github.com/pytorch-labs/gpt-fast}} is a simple yet performant PyTorch-native implementation for transformer text generation. We follow most of its default settings, running benchmarks with a batch size of \texttt{1}.\footnote{This configuration makes the reported tokens per second (tokens/s) equivalent to ``tokens/s/user.'' We set the prompt length to just 1, focusing our measurements on the decoding step in text generation rather than the prefill stage. We also do not use CUDA Graphs due to its incompatibility with \texttt{FLUTE}.} We additionally use \texttt{torch.compile} to optimize the model, which, in early experiments, nearly tripled the throughput of the 16-bit unquantized model.
    \item \texttt{vLLM}~\cite{kwon2023efficient} is a high-throughput and memory-efficient inference and serving engine for LLMs widely used in practice. We benchmarked the latency of processing a single batch of requests, following most of its default settings, but varied the input length, output length, and batch size to assess performance under different conditions.
\end{enumerate}

For the 70B model, the unquantized model does not fit into a single GPU. Consequently, we apply tensor parallelism
across \texttt{4xA6000} or \texttt{2xA100} GPUs. Since the quantized model fits into a single GPU, we report two sets of numbers (single- and multi-GPUs) to represent different use cases.

\input{tables/llama3-results-summary}

\begin{figure*}[t]
\centering
\vspace{-2mm}
\includegraphics[width=.99\textwidth]{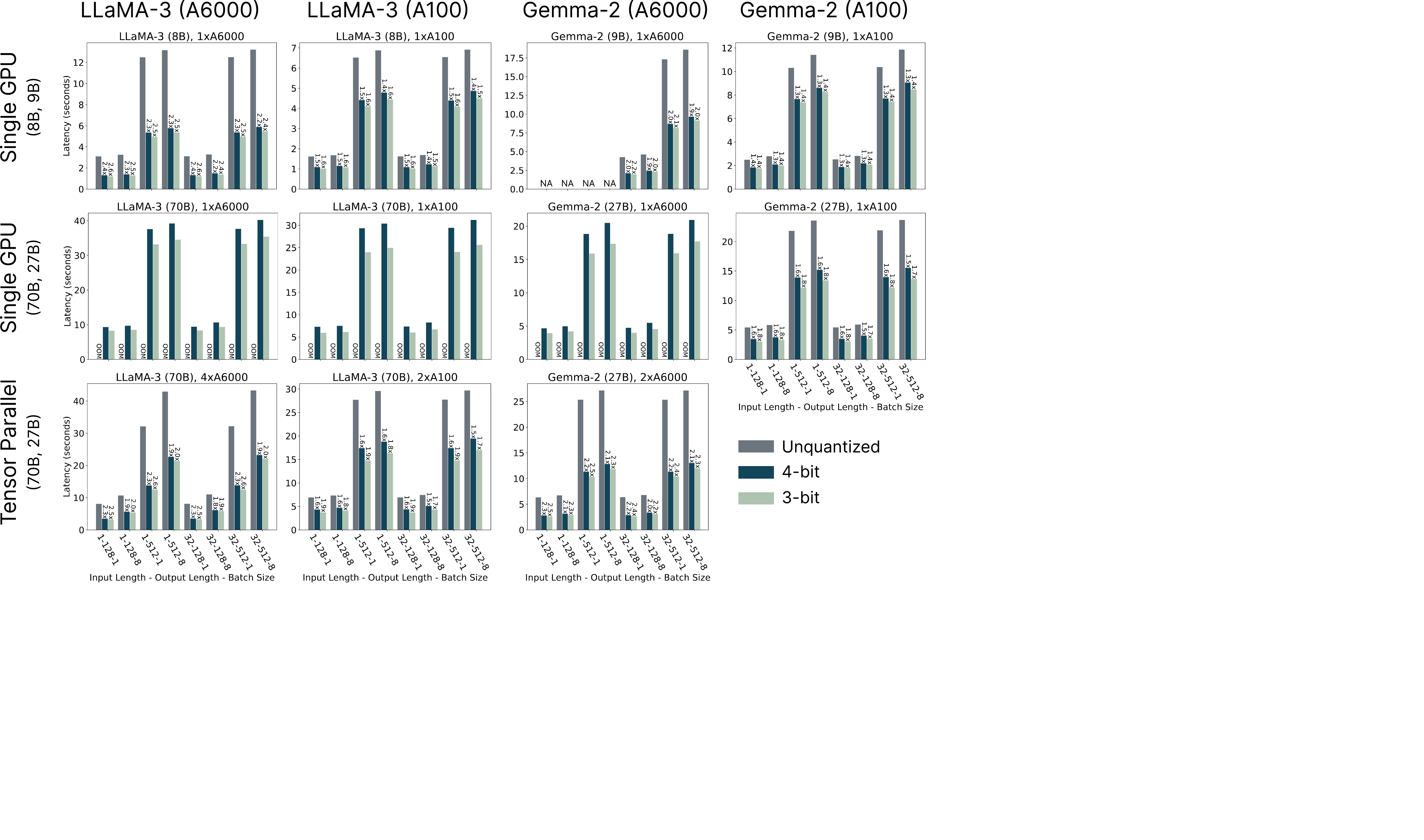}
\caption{
End-to-end latency benchmark for processing a single batch of requests using \texttt{vLLM}. We evaluated LLaMA-3 (8B and 70B) and Gemma-2 (9B and 27B) models with various configurations, including different bits, model sizes, number of GPUs, input lengths, output lengths, and batch sizes. The models were quantized using a group size of \texttt{64} to achieve a good balance between quality and speed.}
\label{fig:vllm}
\vspace{-3mm}
\end{figure*}

\begin{figure}[t]
\centering
\includegraphics[width=.99\linewidth]{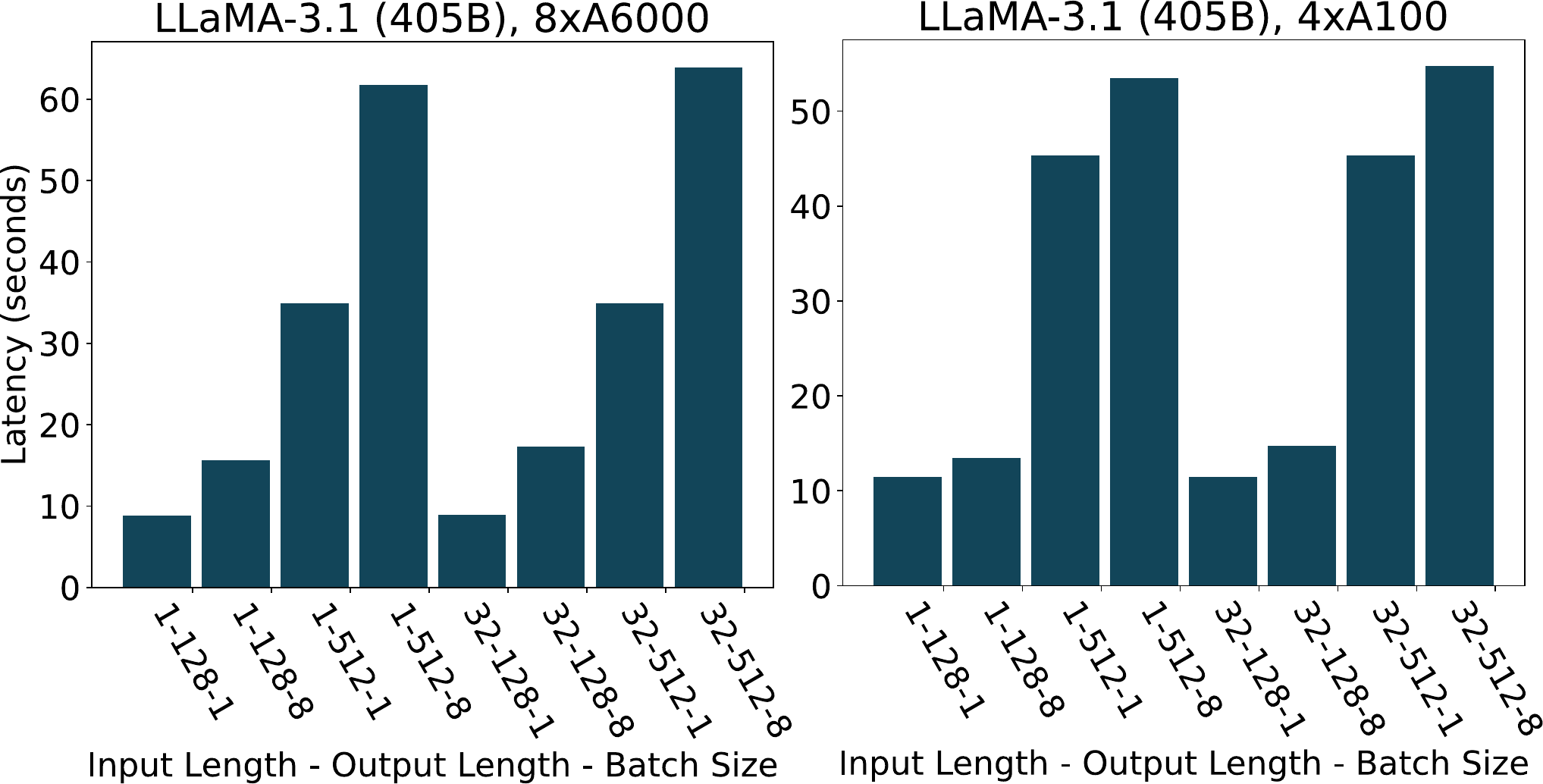}
\vspace{-5mm}
\caption{
End-to-end \texttt{vLLM} latency benchmark for processing a single batch of requests using LLaMA-3.1 (405B, \texttt{W4G64}).}
\label{fig:vllm-llama31-405b}
\vspace{-5mm}
\end{figure}

\vspace{-2mm}
\paragraph{Results.} We first compare our ``learned NF quantization'' approach against  standard 4- and 3-bit setting with group size \texttt{128} against other quantization methods. The results are shown in Table~\ref{tab:ptq_lm3_summary}, where we find that this variant of LUT quantization improves upon ordinary NF quantization and compares favorably against existing baselines. See Tables~\ref{tab:ptq_lm38} and \ref{tab:ptq_lm370} of the appendix for the full results. {We also find that combining NF with AWQ~\cite{lin2023awq} to be beneficial, although a learned NF+AWQ did not help.} (However we emphasize that the quantization method itself is not the main contribution of the present work.) We next exploit the flexibility of \texttt{FLUTE} and conduct end-to-end experiments with various bit- and group-size settings. This is shown in Table~\ref{tab:ptq_lm3_group_sizes}.  With small enough group sizes, our approach is able to almost approach the 16 bit baseline in terms of WikiText2 perplexity.\footnote{Note that the WikiText-2 validation data is different from the calibration data.} We are able to observe meaningful speedups even in the end-to-end case over an optimized baseline.

Finally, we evaluated end-to-end latency using the popular model service framework \texttt{vLLM}~\cite{kwon2023efficient}. Based on our earlier experiments, we selected a group size of \texttt{64}, which strikes a good balance between quality and speed. We conducted experiments across various configurations, including bit precision, model sizes, number of GPUs, input lengths, output lengths, and batch sizes. Additionally, we conducted experiments with the newly released Gemma-2 models (9B and 27B). For the largest open-sourced Gemma-2 27B model, which fits into a \texttt{2xA6000} and \texttt{1xA100} setup, we adjusted the tensor parallelism settings accordingly. The results, presented in Fig.~\ref{fig:vllm}, further showcase the end-to-end performance of the kernel.

To demonstrate the scalability of our approach, we evaluated FLUTE on the LLaMA-3.1 (405B) model~\cite{dubey2024llama}, shown in Fig.~\ref{fig:vllm-llama31-405b}. It is worth noting that, without quantization, the 405B model’s parameters alone would require multiple GPU nodes. However, with \texttt{FLUTE}, we were able to perform inference on a single node, highlighting the efficiency and scalability of our solution.

\vspace{-2mm}
\section{Discussion and Conclusion}
\vspace{-2mm}
Early work on LLM quantization generally worked with uniform quantization methods \cite{frantar-gptq,dettmers2022gpt3,xiao2022smoothquant}. More recent work has shown the benefits of LUT-quantization, both from PTQ \cite{kim2023squeezellm} and finetuning \cite{dettmers2024qlora} perspectives. Insofar as lookup tables can represent flexible quantization functions, our hope is that \texttt{FLUTE} can enable researchers and practitioners to explore new quantization algorithms that can learn better lookup tables \cite{yamamoto2021learnable,cardinaux2020iteratively,wang2022learnable}. For example, recent work has found that codebook-based quantization schemes---which generalize lookup tables to vector-valued values---can enable even lower-bit (e.g., 2-bit) LLM quantization without significant performance degradations \cite{tseng2024quip,egiazarian2024extreme}. We anticipate that ideas from this work can aid in developing kernels for such methods.

Algorithmic considerations aside, one of the main challenges in developing fused quantized matrix multiplication kernels stems from the lack of hardware support for ``mixed-type'' instructions, necessitating software-level implementations. Existing Tensor Core instructions support scenarios where the input and output/accumulation data have different types (e.g., compute in \texttt{FP16} and output/accumulate in \texttt{FP32}). However, they do not support cases where the input operands themselves are of different types (e.g., \texttt{FP16} inputs and \texttt{INT4} weights).
As weight-only quantization becomes increasingly common in LLM inference applications,  native support for such instructions in future hardware could be beneficial. Additionally, the lack of in-register dynamic indexing means that developers must devise software solutions. Enhanced hardware acceleration for indexing into small lookup tables could also prove beneficial in the upcoming generations of AI accelerator hardware.

\section{Conclusion}
This work introduces \texttt{FLUTE}, a CUDA kernel designed for fused quantized matrix multiplications to accelerate LLM inference. \texttt{FLUTE} offers flexibility, supporting flexible mappings between quantized and dequantized values through a lookup table, and accommodating a wide range of bit widths and group sizes. We demonstrate its performance through both kernel-level benchmarks and end-to-end evaluations on state-of-the-art  LLMs.

\section*{Limitations}
\texttt{FLUTE} has several limitations. For one, it is mostly optimized for Ampere-generation GPUs, and it does not take advantage of the newer hardware features available in subsequent generations, such as Hopper GPUs (H100). However, the majority of the methods discussed could still be applicable to the newer hardware. For Ampere generation GPUs, the latest tensor cores support performing \texttt{MMA} operations on matrix fragments of shape \texttt{[16,16]x[16,8]}. When the batch size is smaller than $16$, input data needs to be padded within shared memory. Although this padding increases on-chip data movements (between shared memory and registers) and computations, it does not increase data movement between off-chip and on-chip memory, allowing us to achieve speed-ups in memory-bound cases. In such scenarios, switching to SIMT cores could further enhance performance.
\texttt{FLUTE} is designed for memory-bound scenarios such as LLM decoding. Its performance tends to degrade with larger batch sizes, which are more common during training when the workload becomes more compute-bound.
Finally, while \texttt{FLUTE} demonstrates strong performance among kernels that support LUT-based dequantization, its performance on A100s still falls short of the peak performance that kernels specialized for uniformly quantized matrices can achieve. 

\section*{Acknowledgements}
We thank Yijie Bei and Dmytro Ivchenko for helpful discussion. HG was supported by a Microsoft PhD Fellowship. EX acknowledges the support of NGA HM04762010002, NSF IIS1955532, NSF CNS2008248, NIGMS R01GM140467, NSF IIS2123952, NSF DMS2027737, NSF BCS2040381, NSF DMS2112273, NSF IIS2311990, Semiconductor Research Corporation (SRC) AIHW award 2024AH3210, and DARPA ECOLE HR00112390063. This study was additionally supported by  MIT-IBM Watson AI Lab and the MLA@CSAIL initiative.

%% file: tables/group-sizes.tex
\begin{table*}[t]
    \small
    \centering
    \setlength{\tabcolsep}{.95mm}
    {
    \begin{tabular}{l|rrrr|rr|rr|rr|rr|rr} %
        \toprule
        \textbf{Model} & \multicolumn{4}{c|}{\textbf{Configuration}} & \multicolumn{2}{c|}{\textbf{Perplexity}} & \multicolumn{8}{c}{\textbf{Tokens / Second}}\\
         & Bits & Group & Bits / Param & GB & {WikiText2} & {C4} & \multicolumn{2}{c}{\texttt{1xA6000}} & \multicolumn{2}{c}{\texttt{4xA6000}} &  \multicolumn{2}{c}{\texttt{1xA100}} & \multicolumn{2}{c}{\texttt{2xA100}}\\
        
        \midrule
                  LLaMA-3 8B  
             & 16 & N/A & 16.00 & 15.1 & 6.1 & 8.9 & 44.8 &  1.0\texttt{x} & &  & 90.2 & 1.0\texttt{x} & & \\
             \noalign{\vspace{1pt}}
             \cline{2-15}
             \noalign{\vspace{2pt}}

        & 4 & 32  & 4.50 & 5.7 & 6.1 & 9.4  &  91.3 & 2.0\texttt{x} &            &            & 113.7 & 1.3\texttt{x} &            &            \\
        & 4 & 64  & 4.25 & 5.5 & 6.1 & 9.4  &  95.9 & 2.1\texttt{x} &            &            & 119.4 & 1.3\texttt{x} &            &            \\
        & 4 & 128 & 4.13 & 5.4 & 6.2 & 9.5  &  98.1 & 2.2\texttt{x} &            &            & 121.6 & 1.3\texttt{x} &            &            \\
        & 4 & 256 & 4.06 & 5.4 & 6.3 & 9.5  &  99.8 & 2.2\texttt{x} & \multicolumn{2}{c|}{\texttt{-}} & 121.7 & 1.3\texttt{x} & \multicolumn{2}{c}{\texttt{-}} \\
        & 3 & 32  & 3.50 & 4.9 & 6.9 & 11.0 &  91.9 & 2.1\texttt{x} &            &            & 117.7 & 1.3\texttt{x} &            &            \\
        & 3 & 64  & 3.25 & 4.7 & 7.2 & 11.3 & 104.1 & 2.3\texttt{x} &            &            & 128.5 & 1.4\texttt{x} &            &            \\
        & 3 & 128 & 3.13 & 4.6 & 7.5 & 11.7 & 108.1 & 2.4\texttt{x} &            &            & 133.5 & 1.5\texttt{x} &            &            \\
        & 3 & 256 & 3.06 & 4.6 & 7.9 & 12.2 & 110.0 & 2.5\texttt{x} &            &            & 135.5 & 1.5\texttt{x} &            &            \\

        \midrule
        LLaMA-3 70B
          &   16 & N/A & 16.00 & 131.7 & 2.9 & 6.7 & \texttt{OOM} & \texttt{OOM} & 17.2 & 1.0\texttt{x} & \texttt{OOM} & \texttt{OOM} & 19.9 & 1.0\texttt{x} \\
          \noalign{\vspace{1pt}}
          \cline{2-15}
          \noalign{\vspace{2pt}}
        
        & 4 & 32  & 4.50 & 40.1 & 3.0  & 7.0  & 12.6 & \texttt{-}\;\; & 33.0 & 1.9\texttt{x} & 17.4 & \texttt{-}\;\; & 28.3 & 1.4\texttt{x} \\
        & 4 & 64  & 4.25 & 38.1 & 3.0  & 7.1  & 13.5 & \texttt{-}\;\; & 33.1 & 1.9\texttt{x} & 18.0 & \texttt{-}\;\; & 29.5 & 1.5\texttt{x} \\
        & 4 & 128 & 4.13 & 37.1 & 3.1  & 7.2  & 14.7 & \texttt{-}\;\; & 33.1 & 1.9\texttt{x} & 18.6 & \texttt{-}\;\; & 30.3 & 1.5\texttt{x} \\
        & 4 & 256 & 4.06 & 36.6 & 3.5  & 7.8  & 15.2 & \texttt{-}\;\; & 32.9 & 1.9\texttt{x} & 19.0 & \texttt{-}\;\; & 31.0 & 1.6\texttt{x} \\
        & 3 & 32  & 3.50 & 32.1 & 3.9  & 8.0  & 13.3 & \texttt{-}\;\; & 32.8 & 1.9\texttt{x} & 20.0 & \texttt{-}\;\; & 30.8 & 1.5\texttt{x} \\
        & 3 & 64  & 3.25 & 30.1 & 4.1  & 8.4  & 16.3 & \texttt{-}\;\; & 33.3 & 1.9\texttt{x} & 22.4 & \texttt{-}\;\; & 33.8 & 1.7\texttt{x} \\
        & 3 & 128 & 3.13 & 29.1 & 5.2  & 10.1 & 17.7 & \texttt{-}\;\; & 32.7 & 1.9\texttt{x} & 23.9 & \texttt{-}\;\; & 34.5 & 1.7\texttt{x} \\
        & 3 & 256 & 3.06 & 28.6 & 15.4 & 26.4 & 18.6 & \texttt{-}\;\; & 33.6 & 2.0\texttt{x} & 24.9 & \texttt{-}\;\; & 34.8 & 1.7\texttt{x} \\

        \bottomrule
        
    \end{tabular}}
    \caption{
        Perplexity and decoding speed of LLaMA-3 with learned NF quantization using various quantization configurations. Decoding speedup is measured in tokens per second. The unquantized LLaMA-3 70B model requires multiple GPUs with Tensor Parallelism. Therefore, we report the speed with one GPU, and with Tensor Parallelism applied (labeled as \texttt{x4} and \texttt{x2}). For the 8B models, since all models fit into one GPU, we report only single GPU results.}
           \vspace{-4mm}
    \label{tab:ptq_lm3_group_sizes}
\end{table*}

%% file: tables/llama3-results-summary.tex
\begin{table}[t]
    \small
    \centering

    \setlength{\tabcolsep}{.95mm}
    {
    \begin{tabular}{llrrrrrr}
        \toprule
        \multirow{2}{*}{\textbf{\#P}} &
        \multirow{2}{*}{\textbf{Method}} & 

        \multicolumn{2}{c}{\textbf{Wiki PPL} $\downarrow$} & \multicolumn{2}{c}{\textbf{C4 PPL} $\downarrow$} & 
        \multicolumn{2}{c}{\textbf{LLM Eval} $\uparrow$} 
        \\
        \cmidrule(lr){3-4} \cmidrule(lr){5-6} \cmidrule(lr){7-8} 
        ~ & ~ &  4-bit & 3-bit & 4-bit & 3-bit & 4-bit & 3-bit \\ 
        \midrule
        \textit{8B} & \textit{Unquantized} &  \multicolumn{2}{c}{\textit{6.1}} &  \multicolumn{2}{c}{\textit{8.9}} &  \multicolumn{2}{c}{\textit{68.6}}\\      
         & RTN &  6.7 & 9.7 & 12.2 & 16.6 & 67.8 & 58.7 \\
         & GPTQ &  6.5 & 9.6 & 9.4 & 11.7 & 67.8 & 60.6 \\
         & AWQ &  6.6 & 8.2 & 9.4 & 11.5 & 68.2 & 64.8 \\
         & OmniQuant &  6.6 & 8.4 & 10.1 & 13.5 & 68.3 & 62.4 \\
         \cmidrule(lr){2-8}
         & NF & 6.6 & 9.2 & 9.5 & 13.0 & 68.0 & 62.3 \\

         & NF + AWQ & 6.5 & 8.0 & 9.3 & 11.5 & 67.8 & 65.1 \\

                  & NF (learned) &  6.2 & 7.5 & 9.5 & 11.7 & 67.9 & 63.7 \\
        \midrule
                \textit{70B} & \textit{Unquantized} &  \multicolumn{2}{c}{\textit{2.9}} &  \multicolumn{2}{c}{\textit{6.7}} &  \multicolumn{2}{c}{\textit{75.3}}\\    
         & RTN &  3.6 & 11.6 & 7.9 & 17.1 & 74.0 & 65.3 \\
         & GPTQ &  3.4 & 5.3 & 7.0 & 8.6 & 74.8 & 71.3 \\
         & AWQ &  3.3 & 4.7 & 7.0 & 7.9 & 74.8 & 73.7 \\
         & OmniQuant &  3.3 & 5.4 & 7.5 & 9.3 & 74.2 & 70.2 \\

         \cmidrule(lr){2-8}
         & NF &  3.4 & 8.7 & 7.6 & 16.7 & 74.0 & 64.3 \\
                          & NF + AWQ & 3.2 & 4.6 & 6.9 & 7.8 & 75.2 & 73.8 \\
                                  & NF (learned) &  3.1 & 5.2 & 7.2 & 10.1 & 74.4 & 66.4 \\

        \bottomrule
        
    \end{tabular}}
        \caption{Evaluation of post-training quantization on LLaMA3-8B and LLaMA3-70B. The RTN, GPTQ \cite{frantar-gptq}, AWQ \cite{lin2023awq} results are from \citet{chen2024efficientqat}; the rest are from our implementations. All non-NF methods use uniform weight quantization.}
    \label{tab:ptq_lm3_summary}
        \vspace{-4mm}
\end{table}

%% file: appendix.tex
\section{Appendix}

Please see Algorithm~\ref{alg:flute} for a detailed version of the algorithm, and Tables~\ref{tab:ptq_lm38},~\ref{tab:ptq_lm370} for detailed experimental results.

\input{tables/algorithm}

\input{tables/llama3-8b}

\input{tables/llama3-70b}

%% file: tables/algorithm.tex
\algblock{ParFor}{EndParFor}
\algrenewtext{ParFor}[1]{\algorithmicparfor\ #1\ \algorithmicpardo}
\algrenewtext{EndParFor}{\algorithmicendparfor}

\begin{figure*}[!t]
\begin{minipage}{0.95\textwidth}
\begin{algorithm}[H]
\ttfamily
\scriptsize
\caption{$\texttt{FLUTE}$}
\label{alg:flute}
\begin{algorithmic}
\Require
$\rmX^{\text{g}}$: inputs in HBM \\
\quad\quad\quad\
$\rmQ^{\text{g}}$: quantized weight in HBM \\
\quad\quad\quad\,
$\rmS^{\text{g}}$: scales in HBM \\
\quad\quad\quad\,
$\rmT^{\text{g}}$: lookup table in HBM \\
\quad\quad\quad\,
$\rmY^{\text{g}}$: outputs in HBM \\
\quad\quad\quad\,
$\mathbb{S}_\text{semaphore}$: scratch space for semaphore in HBM \\
\quad\quad\quad\,
$\mathbb{S}_\text{partials}$: scratch space for partials in HBM \\
\State {\color{gray} \# --------------- Offline Preprocessing and Host-side Code ---------------}
\State $\rmQ^\text{g}_1, \rmQ^\text{g}_2  \gets \texttt{reorder\_and\_split}(\rmQ^\text{g})$ \Comment{Section~\ref{ssec:offline}}
\State $\rmT^\text{g}_v \gets \texttt{make\_vectorized\_LUT}(\rmT^\text{g})$ \Comment{Section~\ref{ssec:bandwidth}}
\State $\texttt{tile\_scheduler} \gets \texttt{StreamKTileScheduler}(\text{N}_{\text{SMs}})$ \Comment{Section~\ref{ssec:streamk}}
\State $\text{N}_{\text{blocks}} \gets \texttt{tile\_scheduler.num\_blocks()}$ \Comment{launching blocks proportional to SMs}
\State
\State {\color{gray} \# --------------- Kernel Launches ---------------}
\ParFor{block index b $\gets$ 1 to $\text{N}_{\texttt{blocks}}$} 
    \State \texttt{tile\_scheduler.initialize(b)} \Comment{partitioning works for this block}
    \State $\texttt{allocate}^\text{s} (\rmX^\text{s}, \rmQ_1^\text{s}, \rmQ_2^\text{s}, \rmS^\text{s}, \rmT^\text{s}_v)$ \Comment{allocate circular buffer in shared memory}
    \State $\texttt{allocate}^{\texttt{r}} (\rmX^\text{r}, \rmQ_\text{1}^\text{r}, \rmQ_\text{2}^\text{r}, \rmQ_\text{1+2}^\text{r}, \rmS^\text{r}, \widehat{\rmW}^\text{r}, \rmY^\text{r}, \widehat{\rmY}^\text{r})$ \Comment{allocate fragments in register}
    \State $\text{i}^\text{s}_{\text{read}} \gets 0, \text{i}^\text{s}_{\text{write}} \gets 0$ \Comment{shared memory pipeline index}
    \State $\text{i}^\text{r}_{\text{current}} \gets 0, \text{i}^\text{r}_{\text{next}} \gets 0$ \Comment{register pipeline index}
    \State
    \State {\color{gray} \# --------------- Global Memory to Shared Memory Prefetch ---------------}
    \State $\texttt{copy}^{\texttt{g}\rightarrow\texttt{s}}(\rmT^\text{g}_v, \rmT_v^{s})$
    \For{prefetch stage $\text{t}_{\text{prefetch}} \gets 0$ to $\text{N}_{\text{stages} - 1}$}
        \State $\text{i}^\text{g}_{\text{read}} \gets \texttt{tile\_scheduler.get\_tile\_index}()$
        \State $\texttt{copy}^{\texttt{g}\rightarrow\texttt{s}}(\rmX^\text{g}[\text{b}, \text{:},\text{i}^\text{g}_{\text{read}}], \rmX^{s} [\text{:}, \text{i}^\text{s}_{\text{write}}])$ {\color{gray} \quad \# and the same for $\rmQ_1, \rmQ_2, \rmS$}
        \State $\text{i}^\text{s}_{\text{write}} = (\text{i}^\text{s}_{\text{write}} + 1) \operatorname{mod} \text{N}_{\text{stages}}$
        \State $\texttt{tile\_scheduler.step}()$
    \EndFor
    \State
    \State {\color{gray} \# --------------- Shared to Registers Prefetch ---------------}
    \State \texttt{wait\_for\_one\_tile()}
    \State $\texttt{copy}^{\texttt{s}\rightarrow\texttt{r}}(\rmX^s [\text{i}^\text{r}_{\text{current}},\text{i}^\text{s}_{\text{read}}], \rmX^\text{r} [\text{i}^\text{r}_{\text{current}}])$ {\color{gray} \quad \# and the same for $\rmQ_1, \rmQ_2, \rmS$}
    \State
    \State {\color{gray} \# --------------- Pipelined Main Loop ---------------}
    \State $\rmY^\text{r} \gets \mathbf{0}$  %
    \While{$\neg\texttt{tile\_scheduler.done}()$}
        \For{fragment index $\text{t}_\text{fragment} \gets $ 1 to $\text{N}_{\text{fragments}}$}
            \If{$\text{t}_\text{fragment}$ == $\text{N}_{\text{fragments}}$}
                \State \texttt{wait\_for\_one\_tile()}  \Comment{Wait until the next prefetched tile}
                \State $\text{i}_{\text{read}} = (\text{i}_{\text{read}} + \text{1}) \operatorname{mod} \text{N}_{\text{stages}}$  %
            \EndIf
            \State $\text{i}^\text{r}_{\text{next}} \gets \text{i}^\text{r}_{\text{curr}}$ + 1 \Comment{overlap MMA with next register load}
            \State $\texttt{copy}^{\texttt{s}\rightarrow\texttt{r}}(\rmX^\text{s} [\text{i}^\text{r}_\text{next},\text{i}^\text{s}_{\text{read}}], \rmX^\text{r} [\text{i}^\text{r}_\text{next}])$ {\color{gray} \quad \# and the same for $\rmQ_1, \rmQ_2, \rmS$}
           \If{$\text{t}_\text{fragment}$ == 0}
                \State $\text{i}^\text{g}_{\text{read}} \gets \texttt{tile\_scheduler.get\_tile\_index}()$ \Comment{overlap MMA with SMEM load}
                \State $\texttt{copy}^{\texttt{g}\rightarrow\texttt{s}}(\rmX^\text{g} [\text{b}, \text{:}, \text{i}^\text{g}_{\text{read}}], \rmX^{\text{s}} [ \text{:}, \text{i}^\text{s}_{\text{write}}])$ {\color{gray} \quad \# and the same for $\rmQ_1, \rmQ_2, \rmS$}
                \State $\text{i}^\text{s}_{\text{write}} = (\text{i}^\text{s}_{\text{write}} + 1) \operatorname{mod} \text{N}_{\text{stages}}$
                \State $\texttt{tile\_scheduler.step}()$
            \EndIf
            \State $\rmQ^\text{r}_\text{1+2} \gets \text{combine}(\rmQ^\text{r}_\text{1}, \rmQ^\text{r}_\text{2})$ \Comment{Section~\ref{ssec:offline}}
            \State $\widehat{\rmW}^\text{r} \gets \text{vectorized\_dequantization}(\rmQ^\text{r}_\text{1+2}, \rmS^\text{r}, \rmT^\text{s})$ \Comment{Section~\ref{ssec:bandwidth}}
            \State $\rmY^{\text{r}} \gets \text{tensor\_core\_mma}(\rmY^{\text{r}}, \rmX^{\text{r}}, \widehat{\rmW}^{\text{r}})$
        \EndFor

        \State
        \State {\color{gray} \# --------------- StreamK Fixup (partial sums reductions) ---------------}
        \If{tile\_scheduler.end\_of\_output\_tile()}
            \State $\text{i}_\text{fixup} \gets \text{tile\_scheduler.get\_fixup\_index()}$
            \State $\widehat{\rmY}^\text{r} \gets \text{to\_fp16}(\rmY^\text{r})$ \Comment{Section~\ref{ssec:streamk}}
            \If{$\neg$tile\_scheduler.finished\_output\_tile()}  \Comment{share partial sums through scratch}
                \State accumulate\_and\_store\_partials($\widehat{\rmY}^\text{r}, \mathbb{S}_\text{partials}[\text{i}_\text{fixup}]$)
                \State signal($\mathbb{S}_\text{semaphore}[\text{i}_\text{fixup}]$)
            \Else
                \If{$\neg$tile\_scheduler.started\_output\_tile()} \Comment{aggregate partial sums}
                    \State wait($\mathbb{S}_\text{semaphore}[\text{i}_\text{fixup}]$)
                    \State $\widehat{\rmY}^\text{r} \gets \widehat{\rmY}^\text{r}$ + load\_partials($\mathbb{S}_\text{partials}[\text{i}_\text{fixup}]$)
                \EndIf
                \State $\text{i}_\text{output} \gets \text{tile\_scheduler.get\_output\_tile\_index()}$
                \State epilogue($\widehat{\rmY}^\text{r}, \rmY^\text{g}[\text{i}_\text{output}]$) \Comment{Write output tile from Registers to HBM}
            \EndIf
        \EndIf
    \EndWhile
\EndParFor
\end{algorithmic}
\end{algorithm}%
\end{minipage}
\vspace{-4mm}
\end{figure*}

%% file: tables/llama3-8b.tex
\begin{table*}[t]
    \small
    \centering
    \setlength{\tabcolsep}{.95mm}
    {
    \begin{tabular}{lllrrrrrrrr}
        \toprule
        \multirow{2}{*}{\textbf{Method}} & 
        \multirow{2}{*}{\textbf{Bits}} & 
        \multirow{2}{*}{\textbf{Group}}&
        \multicolumn{2}{c}{\textbf{PPL$\downarrow$}} & 
        \multicolumn{6}{c}{\textbf{LLM Eval$\uparrow$}} 
        \\
        \cmidrule(lr){4-5} \cmidrule(lr){6-11} 
        ~ & ~ & ~ & \textbf{WikiText2} & \textbf{C4} & \textbf{PIQA} &\textbf{ARC-e} & \textbf{ARC-c} &\textbf{HellaSwag} & \textbf{Wino} & \textbf{Avg.} \\ 
        \midrule
        \textit{Unquantized} & 16 & N/A & 6.1 & 8.9 & 79.6 & 80.1 & 50.4 & 60.2 & 72.6 & 68.6 \\
        \midrule
        
        \multirow{2}{*}{RTN} 
         & 4 & 128 & 6.7 & 9.7
         & 79.1 & 79.3 & 48.4 & 59.0 & 73.2 & 67.8 \\
         & 3 & 128 & 12.2 & 16.6
         & 74.2 & 65.4 & 36.7 & 50.9 & 66.5 & 58.7 \\
        \midrule
        
        \multirow{2}{*}{GPTQ} 
         &  4  &  128  &  6.5  &  9.4
         & 78.9 & 79.4 & 47.7 & 59.2 & 73.7 & 67.8  \\
         &  3  &  128  &  9.6  &  11.7 
         & 73.8 & 65.2 & 37.8 & 55.1 & 70.8 & 60.6  \\
        \midrule

        \multirow{2}{*}{AWQ} 
        & 4 & 128 & 6.6 & 9.4 & 79.2 & 79.6 & 50.0 & 59.4 & 73.0 & 68.2 \\
        & 3 & 128 & 8.2 & 11.5 & 77.7 & 75.8 & 44.2 & 55.4 & 71.0 & 64.8 \\
        \midrule

        \multirow{2}{*}{OmniQuant} 
        & 4 & 128 & 6.6 & 10.1 & 79.1 & 80.0 & 49.7 & 59.4 & 73.2 & 68.3 \\
        & 3 & 128 & 8.4 & 13.5 & 76.4 & 70.0 & 40.9 & 55.1 & 69.5 & 62.4 \\
        \midrule

        \multirow{2}{*}{NormalFloat} 
        & 4 & 128 & 6.6 & 9.5 & 78.6 & 79.6 & 49.6 & 59.0 & 73.5 & 68.0 \\
        & 3 & 128 & 9.2 &  13.0 & 75.4 & 72.0 & 40.5 & 54.4 & 69.4 & 62.3 \\\midrule

        NormalFloat & 4 & 128 & 6.2 & 9.5 & 79.0 & 79.6 & 49.0 & 59.4 & 72.6 & 67.9 \\
        learned $\sigma$ & 3 & 128 & 7.5 &  11.7 & 77.1 & 74.1 & 41.7 & 55.8 & 69.7 & 63.7 \\\midrule

        NormalFloat & 4 & 128 & 6.5 & 9.3 & 79.6 & 78.0 & 48.5 & 59.0 & 73.8 & 67.8 \\
         + AWQ & 3 & 128 & 8.0 & 11.5 & 77.0 & 75.5 & 44.6 & 55.9 & 72.3 & 65.1 \\

        \bottomrule
        
    \end{tabular}}
    \caption{Detailed evaluation of post-training quantization on LLaMA3-8B. The RTN, GPTQ \cite{frantar-gptq}, AWQ \cite{lin2023awq} results are from \citet{chen2024efficientqat}; the rest are from our implementations. All non-NormalFloat methods use uniform weight quantization.}
    \label{tab:ptq_lm38}
\end{table*}

%% file: tables/llama3-70b.tex
\begin{table*}[t]
    \small
    \centering
    \setlength{\tabcolsep}{.95mm}
    {
    \begin{tabular}{lllrrrrrrrr}
        \toprule
        \multirow{2}{*}{\textbf{Method}} & 
        \multirow{2}{*}{\textbf{Bits}} & 
        \multirow{2}{*}{\textbf{Group}}&
        \multicolumn{2}{c}{\textbf{PPL$\downarrow$}} & 
        \multicolumn{6}{c}{\textbf{LLM Eval$\uparrow$}} 
        \\
        \cmidrule(lr){4-5} \cmidrule(lr){6-11} 
        ~ & ~ & ~ & \textbf{WikiText2} & \textbf{C4} & \textbf{PIQA} &\textbf{ARC-e} & \textbf{ARC-c} &\textbf{HellaSwag} & \textbf{Wino} & \textbf{Avg.} \\ 
        \midrule
        \textit{Unquantized} & 16 & N/A & 2.9 & 6.7 & 82.4 & 87.0 & 60.4 & 66.4 & 80.5 & 75.3 \\
        \midrule
        
        \multirow{2}{*}{RTN} 
         & 4 & 128 & 3.6 & 7.9 & 82.3&85.7&57.3&65.8&78.8&74.0\\
         & 3 & 128 & 11.6 & 17.1 & 79.1 & 78.7 & 48.5 & 54.2 & 65.9 & 65.3\\
        \midrule
        
        \multirow{2}{*}{GPTQ} 
         & 4 & 128 & 3.4 & 7.0 &82.3&85.7&59.0&66.1&80.5&74.8\\
         & 3 & 128 & 5.3 & 8.6 &80.6&82.1&53.0&62.6&78.1&71.3 \\
        \midrule

        \multirow{2}{*}{AWQ} 
        & 4 & 128 & 3.3 & 7.0 & 82.2 & 86.4 & 59.1 & 65.8 & 80.4 & 74.8 \\
        & 3 & 128 & 4.7 & 7.9 & 82.3 & 84.5 & 58.4 & 64.3 & 78.9 & 73.7 \\
        \midrule

        \multirow{2}{*}{OmniQuant} 
        & 4 & 128 & 3.3 & 7.5 & 82.0 & 85.6 & 58.0 & 66.0 & 79.6 & 74.2 \\
        & 3 & 128 & 5.4 & 9.3 & 80.8 & 80.6 & 50.9 & 63.7 & 75.2 & 70.2 \\
        \midrule

        \multirow{2}{*}{NormalFloat} 
        & 4 & 128 & 3.4 &  7.6 & 82.0 & 85.6 & 56.7 & 66.1 & 79.5 & 74.0 \\
        & 3 & 128 & 8.7 & 16.7 & 76.6 & 76.9 & 42.7 & 55.8 & 69.3 & 64.3 \\\midrule

        NormalFloat & 4 & 128 & 3.1 &  7.2 & 82.3 & 85.7 & 58.2 & 66.4 & 79.6 & 74.4 \\
        learned $\sigma$ & 3 & 128 & 5.2 & 10.1 & 77.3 & 76.7 & 44.3 & 62.4 & 71.2 & 66.4 \\\midrule

        NormalFloat & 4 & 128 & 3.2 & 6.9 & 82.6 & 86.8 & 60.1 & 65.9 & 80.5
 & 75.2 \\
         + AWQ & 3 & 128 & 4.6 & 7.8 & 81.4 & 85.3 & 58.5 & 64.6 & 79.2 & 73.8 \\

        \bottomrule
        
    \end{tabular}}
    \caption{Detailed evaluation of post-training quantization on LLaMA3-70B. The RTN, GPTQ \cite{frantar-gptq}, AWQ \cite{lin2023awq} results are from \citet{chen2024efficientqat}; the rest are from our implementations. All non-NormalFloat methods use uniform weight quantization.}
    \label{tab:ptq_lm370}
\end{table*}